\documentclass[sigconf]{acmart}
\usepackage{multirow}
\usepackage{amsmath}
\usepackage{caption}
\usepackage{booktabs}
\usepackage{subcaption}
\usepackage{caption}
\usepackage{multicol} 
\usepackage[font=small, labelfont=bf, textfont=normalfont]{caption} 
\AtBeginDocument{%
  }

\begin{document}

\title{Optimizing TinyML: The Impact of Reduced Data Acquisition Rates for Time Series Classification on Microcontrollers}

\author{ Riya Samanta, Bidyut Saha, Soumya K. Ghosh, and Ram Babu Roy}
\affiliation{%
  \institution{Indian Institute of Technology Kharagpur \country{India}}}
\email{ riya.samanta@iitkgp.ac.in,bidyutsaha@kgpian.iitkgp.ac.in,  rambabu@see.iitkgp.ac.in ,  skg@cse.iitkgp.ac.in }


\begin{abstract}
Tiny Machine Learning (TinyML) enables efficient, low-cost, and privacy-preserving machine learning inference directly on microcontroller units (MCUs) connected to sensors. Optimizing models for these constrained environments is crucial. This paper investigates how reducing data acquisition rates affects TinyML models for time series classification, focusing on resource-constrained, battery-operated IoT devices. By lowering data sampling frequency, we aim to reduce computational demands—RAM usage, energy consumption, latency, and MAC operations—by approximately fourfold while maintaining similar classification accuracies. Our experiments with six benchmark datasets (UCIHAR, WISDM, PAMAP2, MHEALTH, MIT-BIH, and PTB) showed that reducing data acquisition rates significantly cut energy consumption and computational load, with minimal accuracy loss. For example, a 75\% reduction in acquisition rate for MIT-BIH and PTB datasets led to a 60\% decrease in RAM usage, 75\% reduction in MAC operations, 74\% decrease in latency, and 70\% reduction in energy consumption, without accuracy loss. These results offer valuable insights for deploying efficient TinyML models in constrained environments.
\end{abstract}



\maketitle

\section{Introduction}
TinyML refers to the deployment of machine learning algorithms on resource-constrained devices, such as microcontrollers. These devices have very limited computational power, memory, and energy resources, often measured in the range of hundreds of kilobytes of RAM and flash memory. Therefore, optimizing both the models and the data they process is crucial. One key factor in such optimization is the rate at which data is acquired and processed. This paper investigates how reducing the data acquisition rate impacts the performance of TinyML models for time series classification.

\begin{figure}[ht]
\vspace{-0.15in}
    \centering
    \includegraphics[width=0.9\linewidth]{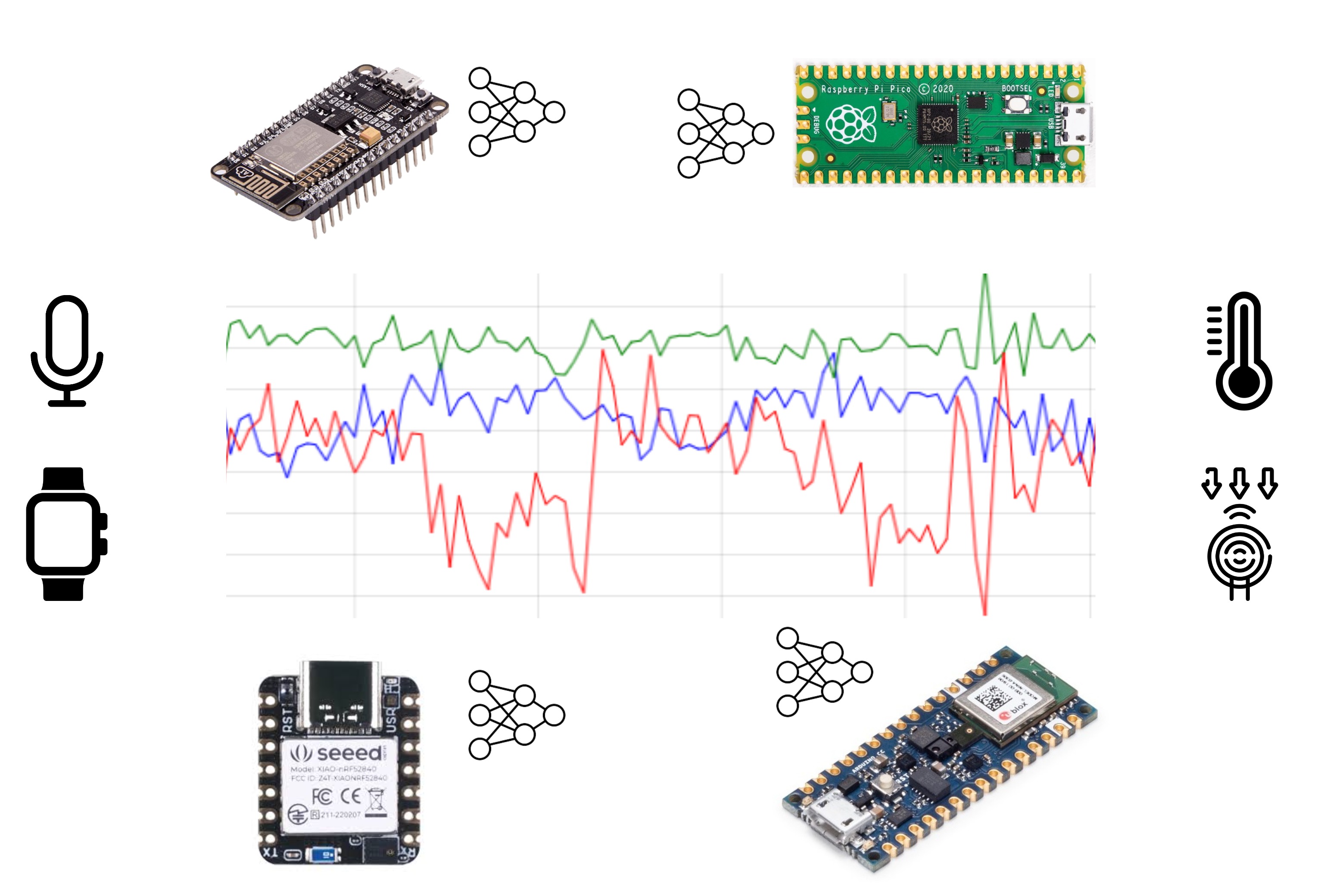}
    \caption{Time series signals from various sensors (e.g., accelerometer, gyroscope, magnetometer, temperature, humidity,  presence sensors etc) are processed by TinyML models running directly on MCUs (such as ESP32, Raspberry Pi Pico, BLE 33 Nano,  NRF52840 etc).}
    \label{fig:yourlabel}
    \vspace{-0.1in}
\end{figure}

\vspace{-0.15in}
\section{Background }
\subsection{TinyML and Resource Constraints}
TinyML has emerged as a field dedicated to bringing machine learning capabilities to MCUs connected with sensors where they are deployed, operating with limited resources. The primary challenge in this field is to develop efficient models that operate within these constraints without sacrificing performance. This involves creating models that can function effectively on devices with restricted computational power, memory, and energy resources 

\subsection{Time Series Classification}
Time series classification involves categorizing data points collected sequentially over time. Applications of time series classification include human activity recognition \cite{saha2023bandx, saha2023tinyml, saha2023wrist, saha2024personalized} from wearable sensors, health monitoring, and predictive maintenance. Benchmark datasets such as UCI HAR, WISDM, PAMAP2, and MHEALTH are commonly used in these areas. For ECG classification, datasets like MIT-BIH and PTB are frequently employed.

\subsection{Data Acquisition Rate}
The data acquisition rate, or sampling rate, determines how frequently data is collected. Higher rates provide more detailed information but require more processing power and memory, while lower rates conserve resources but may miss important details. The balance between data granularity and resource consumption is critical for optimizing performance on resource-constrained devices.

Previous studies have explored various strategies for optimizing TinyML models, including model compression, quantization, and hardware-specific optimizations. However, the impact of reducing data acquisition rates on TinyML model performance has not been extensively studied. This research aims to fill this gap by examining how different data acquisition rates affect the efficiency and effectiveness of TinyML models for time series classification.

\section{Methodology}
\subsection{Datasets}
In our experiments, we utilized six benchmark datasets spanning the domains of lifestyle, healthcare, and human-computer interaction. Human activity recognition directly impacts lifestyle and human-computer interaction domains, and indirectly contributes to healthcare. We employed the UCIHAR, WISDM, PAMAP2, and MHEALTH datasets, and for healthcare applications, we leveraged the MIT-BIH and PTB Diagnostic ECG Database. Details for each dataset are provided below.

\vspace{-0.1in}
\begin{table}[!t]
\centering
\caption{Classification Model Architecture}
\label{tab:model_architecture}
\vspace{-0.1in}
\resizebox{0.5\textwidth}{!}{
\begin{tabular}{|l|c|c|c|c|}
\hline
\textbf{Type} & \textbf{Stride} & \textbf{Size} & \textbf{Filters/Neurons} & \textbf{Activation} \\
\hline
Input & - & - & - & - \\
SeparableConv1D & 1 & 3 & 32 & ReLU \\

MaxPool1D & 2 & 2 & - & - \\

SeparableConv1D & 1 & 3 & 48 & ReLU \\

MaxPool1D & 2 & 2 & - & - \\

SeparableConv1D & 1 & 3 & 72 & ReLU \\

MaxPool1D & 2 & 2 & - & - \\

GlobalAveragePooling1D & - & - & - & - \\

Dense & - & - & 72 & ReLU \\

Dropout (0.25) & - & - & - & - \\

Dense & - & - & \#classes & Softmax \\
\hline
\end{tabular}%
}
\vspace{-0.2in}
\end{table}

\subsubsection{UCIHAR}
The UCI Human Activity Recognition \cite{misc_human_activity_recognition_using_smartphones_240} dataset captures daily activities using a waist-mounted smartphone with inertial sensors. It includes recordings of six activities (WALKING, WALKING-UPSTAIRS, WALKING-DOWNSTAIRS, SITTING, STANDING, LAYING) from 30 subjects aged 19-48. The dataset includes 3-axis linear acceleration and 3-axis angular velocity data captured at 50Hz, processed into fixed-width sliding windows of 2.56 seconds with 50\% overlap. Features were extracted from both time and frequency domains after noise filtering and gravitational separation using a low-pass filter with a cutoff frequency of 0.3 Hz.

\subsubsection{WISDM}
The WISDM \cite{misc_wisdm_smartphone_and_smartwatch_activity_and_biometrics_dataset__507} dataset contains accelerometer and gyroscope time-series sensor data collected from a smartphone and smartwatch as 51 subjects perform 18 activities for 3 minutes each. The raw sensor data, sampled at 20Hz, is gathered from both devices. We used data from the smartwatch's accelerometer and gyroscope with a 10 second window and 50\% overlap to generate features, focusing on six activities: WALKING, JOGGING, TYPING, WRITING, STAIRS, and BRUSHING-TEETH .

\subsubsection{PAMAP2}
The PAMAP2 \cite{misc_pamap2_physical_activity_monitoring_231} dataset captures data from 18 physical activities performed by 9 subjects using 3 IMU sensors and a heart rate monitor. We consider 6 activities: WALKING, RUNNING, CYCLING, COMPUTER-WORK, CAR-DRIVING, and ROPE-JUMPING. Data from the wrist IMU is used,  sampled at 100Hz. We applied a sliding window of 5.12 seconds with 50\% overlap for processing, focusing on accelerometer and gyroscope readings .

\subsubsection{MHEALTH}
The MHEALTH \cite{misc_mhealth_319} dataset comprises body motion and vital signs recordings from ten volunteers of diverse profiles performing various physical activities at 50Hz. We utilized only the sensors placed on the subjects' right wrists, focusing on five activities: WALKING, CLIMBING-STAIRS, CYCLING, JOGGING, and RUNNING. For feature extraction, we used a 5-second sliding window with a 50\% overlap.

\subsubsection{MIT-BIH}
The MIT-BIH Arrhythmia  Database \cite{moody2001impact} contains 48 half-hour excerpts of two-channel ambulatory ECG recordings from 47 subjects studied between 1975 and 1979. These recordings were digitized at 360 samples per second per channel. The database includes approximately 110,000 beats annotated by multiple cardiologists, covering both common and clinically significant arrhythmias. We used ECG lead II data resampled to 125Hz and classified beats into five categories (N, S, V, F, Q). For preprocessing the ECG beats, we employed a sliding window of 1.5 seconds and followed the pipeline described by Kachuee et al. \cite{kachuee2018ecg}.

\subsubsection{PTB Diagnostic ECG Database}

The PTB Diagnostic ECG Database \cite{bousseljot1995nutzung} comprises ECG records from 290 subjects, including 148 diagnosed with MI, 52 healthy controls, and others with various diseases. Each record includes ECG signals from 12 leads sampled at 1000Hz. We focused on ECG lead II, resampled to 125Hz, and analyzed the MI and healthy control categories using a 1.5-second sliding window. For preprocessing the ECG beats, we followed the pipeline detailed by Kachuee et al. \cite{kachuee2018ecg}.

\subsection{Data Preprocessing}
Each dataset was preprocessed to create multiple versions with different sampling rates. We produced downsampled versions at reduction rates of 0\%, 25\%, 50\%, and 75\% of the original frequency. For details, refer to Tables [\ref{ucihar}, \ref{wisdm}, \ref{pamap2}, \ref{mhealth}, \ref{mitbih}, \ref{ptbd}].

\subsection{Model Architecture}
We employed a TinyML-optimized neural network architecture consisting of SeparableConv1D, MaxPooling1D, GlobalAveragePooling1D, and Dense layers. Full architectural details are provided in Table \ref{tab:model_architecture}. The model uses the categorical cross-entropy loss function and the Adam optimizer with a learning rate of 0.001 and a batch size of 32. This model was trained on different sampling rate variations of each dataset using TensorFlow \cite{david2021tensorflow}.

\begin{table}[!t]
\centering
\caption{UCIHAR dataset performance with varying data acquisition reductions. Metrics include accuracy, FLASH, RAM, MACs, time, and energy. Reduction (\%) indicates the decrease in sampling frequency.}
\label{ucihar}
\vspace{-0.1in}

\resizebox{0.5\textwidth}{!}{%
\begin{tabular}{|c|c|c|c|c|c|c|c|}
\hline
\textbf{\begin{tabular}[c]{@{}c@{}}Reduction \\ (\%)\end{tabular}} & \textbf{\begin{tabular}[c]{@{}c@{}}Freq. (Hz),  \\ Input Shape\end{tabular}} & \textbf{Accuracy} & \textbf{\begin{tabular}[c]{@{}c@{}}FLASH \\(KB)\end{tabular}} & \textbf{\begin{tabular}[c]{@{}c@{}}RAM \\(KB)\end{tabular}} & \textbf{\begin{tabular}[c]{@{}c@{}}MACs \\(K)\end{tabular}} & \textbf{\begin{tabular}[c]{@{}c@{}}Latency\\ (ms)\end{tabular}} & \textbf{\begin{tabular}[c]{@{}c@{}}Energy\\ (uJ)\end{tabular}} \\ \hline
0 & 50, (128 $\times$ 9) & 0.928 & 28.4 & 13.6 & 265.6 & 25.0 & 139.5 \\ \hline
25 & 37, (96 $\times$ 9) & 0.927 & 28.5 & 11.6 & 200.6 & 18.9 & 129.4 \\ \hline
50 & 25, (64 $\times$ 9) & 0.929 & 28.6 & 9.5 & 135.6 & 13.2 & 120.7 \\ \hline
75 & 12, (32 $\times$ 9) & 0.904 & 28.6 & 6.25 & 118.83 & 7.69 & 40.88 \\ \hline
\end{tabular}%
}

\end{table}

\begin{table}[!t]
\centering
\caption{WISDM dataset performance with varying data acquisition reductions. Metrics include accuracy, FLASH, RAM, MACs, time, and energy. Reduction (\%) indicates the decrease in sampling frequency.}
\label{wisdm}
\vspace{-0.1in}

\resizebox{0.5\textwidth}{!}{%
\begin{tabular}{|c|c|c|c|c|c|c|c|}
\hline
\textbf{\begin{tabular}[c]{@{}c@{}}Reduction \\ (\%)\end{tabular}} & \textbf{\begin{tabular}[c]{@{}c@{}}Freq. (Hz),  \\ Input Shape\end{tabular}} & \textbf{Accuracy} & \textbf{\begin{tabular}[c]{@{}c@{}}FLASH \\(KB) \end{tabular}} & \textbf{\begin{tabular}[c]{@{}c@{}}RAM \\(KB) \end{tabular}} & \textbf{\begin{tabular}[c]{@{}c@{}}MACs \\(K) \end{tabular}} & \textbf{\begin{tabular}[c]{@{}c@{}}Latency\\ (ms)\end{tabular}} & \textbf{\begin{tabular}[c]{@{}c@{}}Energy\\ (uJ)\end{tabular}} \\ \hline
0 & 20, (200 $\times$ 6) & 0.968  & 27.3 &     18.1 &	390.8 &		37.6&	295.8 \\
25 & 15, (150 $\times$ 6) & 0.949  & 27.6 &	 14.9 &	292.7 &		28.4&	162.2 \\
50 & 10, (100 $\times$ 6) & 0.925	& 27.5 &	11.7 &	198.2 &	    19.6&	132.5 \\
75 & 5, (50 $\times$ 6) & 0.895	& 27.5 &	6.46 &	104.21 &    10.5&	51.75 \\
\hline
\end{tabular}%
}
\vspace{-0.1in}

\end{table}

\begin{table}[!t]
\centering
\caption{PAMAP2 dataset performance with varying data acquisition reductions. Metrics include accuracy, FLASH, RAM, MACs, time, and energy. Reduction (\%) indicates the decrease in sampling frequency.}
\label{pamap2}
\vspace{-0.1in}

\resizebox{0.5\textwidth}{!}{%
\begin{tabular}{|c|c|c|c|c|c|c|c|}
\hline
\textbf{\begin{tabular}[c]{@{}c@{}}Reduction \\ (\%)\end{tabular}} & \textbf{\begin{tabular}[c]{@{}c@{}}Freq. (Hz),  \\ Input Shape\end{tabular}} & \textbf{Accuracy} & \textbf{\begin{tabular}[c]{@{}c@{}}FLASH \\(KB) \end{tabular}} & \textbf{\begin{tabular}[c]{@{}c@{}}RAM \\(KB) \end{tabular}} & \textbf{\begin{tabular}[c]{@{}c@{}}MACs \\(K) \end{tabular}} & \textbf{\begin{tabular}[c]{@{}c@{}}Latency\\ (ms)\end{tabular}} & \textbf{\begin{tabular}[c]{@{}c@{}}Energy\\ (uJ)\end{tabular}} \\ \hline
0 & 100, (512 $\times$ 6) & 0.98 &	    28.5 &	38.1 &	991.7 & 94.5 &	1000\\
25 & 75, (384 $\times$ 6) & 0.972 &	28.5 &	29.9 &	745.2 & 70.2 &	675.2  \\
50 & 50, (256 $\times$ 6) & 0.962 & 	28.5 &	21.7 &	498.7 &  48.5 &	467.2 \\
75 & 25, (128 $\times$ 6) & 0.947 &	28.5 &	13.5 &	252.1 & 25.0 &	150.5 \\
\hline
\end{tabular}%
}
\vspace{-0.1in}

\end{table}

\begin{table}[!t]
\centering
\caption{MHEALTH dataset performance with varying data acquisition reductions. Metrics include accuracy, FLASH, RAM, MACs, time, and energy. Reduction (\%) indicates the decrease in sampling frequency.}
\label{mhealth}
\vspace{-0.1in}

\resizebox{0.5\textwidth}{!}{%
\begin{tabular}{|c|c|c|c|c|c|c|c|}
\hline
\textbf{\begin{tabular}[c]{@{}c@{}}Reduction \\ (\%)\end{tabular}} & \textbf{\begin{tabular}[c]{@{}c@{}}Freq. (Hz),  \\ Input Shape\end{tabular}} & \textbf{Accuracy} & \textbf{\begin{tabular}[c]{@{}c@{}}FLASH \\(KB) \end{tabular}} & \textbf{\begin{tabular}[c]{@{}c@{}}RAM \\(KB) \end{tabular}} & \textbf{\begin{tabular}[c]{@{}c@{}}MACs \\(K) \end{tabular}} & \textbf{\begin{tabular}[c]{@{}c@{}}Latency\\ (ms)\end{tabular}} & \textbf{\begin{tabular}[c]{@{}c@{}}Energy\\ (uJ)\end{tabular}} \\ \hline
0 & 50, (250 $\times$ 6) & 0.956 &	28.2 &	21.3 &	485.2 &  46.8 &	    454.9 \\
25 & 35, (187 $\times$ 6) & 0.92 &	    28.2& 	17.3 &	362.2 &   34.1 &	228.1 \\
50 & 25, (125 $\times$ 6) & 0.9051 &   28.3 &	13.4 &	244.6 &   24.7 &     152.1\\
75 & 12, (62 $\times$ 6) & 0.898 &	28.4 & 	9.3 &	123.2 &   12.6 &	124.3\\
\hline
\end{tabular}%
}
\vspace{-0.1in}

\end{table}

\begin{table}[!t]
\centering
\caption{MIT-BIH Arrhythmia Database dataset performance with varying data acquisition reductions. Metrics include accuracy, FLASH, RAM, MACs, time, and energy. Reduction (\%) indicates the decrease in sampling frequency.}
\label{mitbih}
\vspace{-0.1in}

\resizebox{0.5\textwidth}{!}{%
\begin{tabular}{|c|c|c|c|c|c|c|c|}
\hline
\textbf{\begin{tabular}[c]{@{}c@{}}Reduction \\ (\%)\end{tabular}} & \textbf{\begin{tabular}[c]{@{}c@{}}Freq. (Hz),  \\ Input Shape\end{tabular}} & \textbf{Accuracy} & \textbf{\begin{tabular}[c]{@{}c@{}}FLASH \\(KB) \end{tabular}} & \textbf{\begin{tabular}[c]{@{}c@{}}RAM \\(KB) \end{tabular}} & \textbf{\begin{tabular}[c]{@{}c@{}}MACs \\(K) \end{tabular}} & \textbf{\begin{tabular}[c]{@{}c@{}}Latency\\ (ms)\end{tabular}} & \textbf{\begin{tabular}[c]{@{}c@{}}Energy\\ (uJ)\end{tabular}} \\ \hline
0 & 125, (187 $\times$ 1) & 0.981 &	27.1 &	17.2 &	329.5 & 32.8 &	219.1 \\
25 & 93.75, (140$\times$1) & 0.981 &	27.1 &	14.3 &	250.7 & 26.1 &	147.6 \\
50 & 62.5, (93$\times$1) & 0.98 &	    27.1 &	11.2 &	166.7 & 17.9 &	131.2 \\
75 & 31.25, (46 $\times$ 1) & 0.98 &	    27.1 &	6.88 &	84.35 & 8.63 &	65.03\\
\hline
\end{tabular}%
}
\vspace{-0.1in}

\end{table}

\begin{table}[!t]
\centering
\caption{PTB Diagnostic ECG Database dataset performance with varying data acquisition reductions. Metrics include accuracy, FLASH, RAM, MACs, time, and energy. Reduction (\%) indicates the decrease in sampling frequency.}
\label{ptbd}
\vspace{-0.1in}
\resizebox{0.5\textwidth}{!}{%
\begin{tabular}{|c|c|c|c|c|c|c|c|}
\hline
\textbf{\begin{tabular}[c]{@{}c@{}}Reduction \\ (\%)\end{tabular}} & \textbf{\begin{tabular}[c]{@{}c@{}}Freq. (Hz),  \\ Input Shape\end{tabular}} & \textbf{Accuracy} & \textbf{\begin{tabular}[c]{@{}c@{}}FLASH \\(KB) \end{tabular}} & \textbf{\begin{tabular}[c]{@{}c@{}}RAM \\(KB) \end{tabular}} & \textbf{\begin{tabular}[c]{@{}c@{}}MACs \\(K) \end{tabular}} & \textbf{\begin{tabular}[c]{@{}c@{}}Latency\\ (ms)\end{tabular}} & \textbf{\begin{tabular}[c]{@{}c@{}}Energy\\ (uJ)\end{tabular}} \\ \hline
0 & 125, (187$\times$1) & 0.976 &	27.9 &	17.3 &	329.5 &  32.8 &	219.1\\
25 & 93.75, (140$\times$1) & 0.977 &	28.0 &	14.3 &	250.7 &  26.1 &	147.6\\
50 & 62.5, (93$\times$1) & 0.971 &	28.1 &	11.3 &	166.7 &  17.9 &	131.2 \\
75 & 31.25, (46$\times$1) & 0.98 &	    28.1 &	6.88 &	84.35 &  8.63 &	65.03\\
\hline
\end{tabular}%
}
\vspace{-0.1in}
\end{table}

\vspace{-0.1in}
\subsection{Optimization and Deployment on MCUs}
The trained models cannot be directly deployed in resource constrained MCU environments due to tight resource constraints. To address this, we utilize TensorFlow Lite Micro \cite{david2021tensorflow} as a framework for deploying ML models on MCUs. We optimize the model using TensorFlow Lite \cite{tensorflowlite} through pruning and post-training integer quantization, making it compatible with the TensorFlow Lite Micro framework. Additionally, we use the Machine Learning Toolkit (MLTK) \cite{mltk} to estimate the resource footprints of the optimized models in MCU environments. MLTK provides estimates based on the ARM Cortex-M33, running the TensorFlow Lite Micro ARM CMSIS and reference kernels in a basic simulator. All profiling information is estimated, with numbers based on the EFR32xG24 at 78MHz.

\subsection{Experiments}

As previously mentioned, we downsampled each dataset to create four versions at reduced data acquisition rates of 0\%, 25\%, 50\%, and 75\%. For example, the UCIHAR dataset originally has a 50Hz sampling rate with a 2.56-second sliding window, resulting in a data instance shape of 128x9. For a 0\% reduction, the shape remains 128x9. For a 25\% reduction, the new sampling rate is 37Hz with a shape of 96x9. A 50\% reduction results in a 25Hz sampling rate with a shape of 64x9, and a 75\% reduction results in a 12Hz sampling rate with a shape of 32x9. Each of these versions was trained using the classification model detailed in Table \ref{tab:model_architecture}, optimized, and then profiled for different resource footprints using the MLTK library.

\begin{figure*}[!h]
\centering

\begin{multicols}{3}
\subcaptionbox{UCIHAR}{\includegraphics[width=0.9\linewidth]{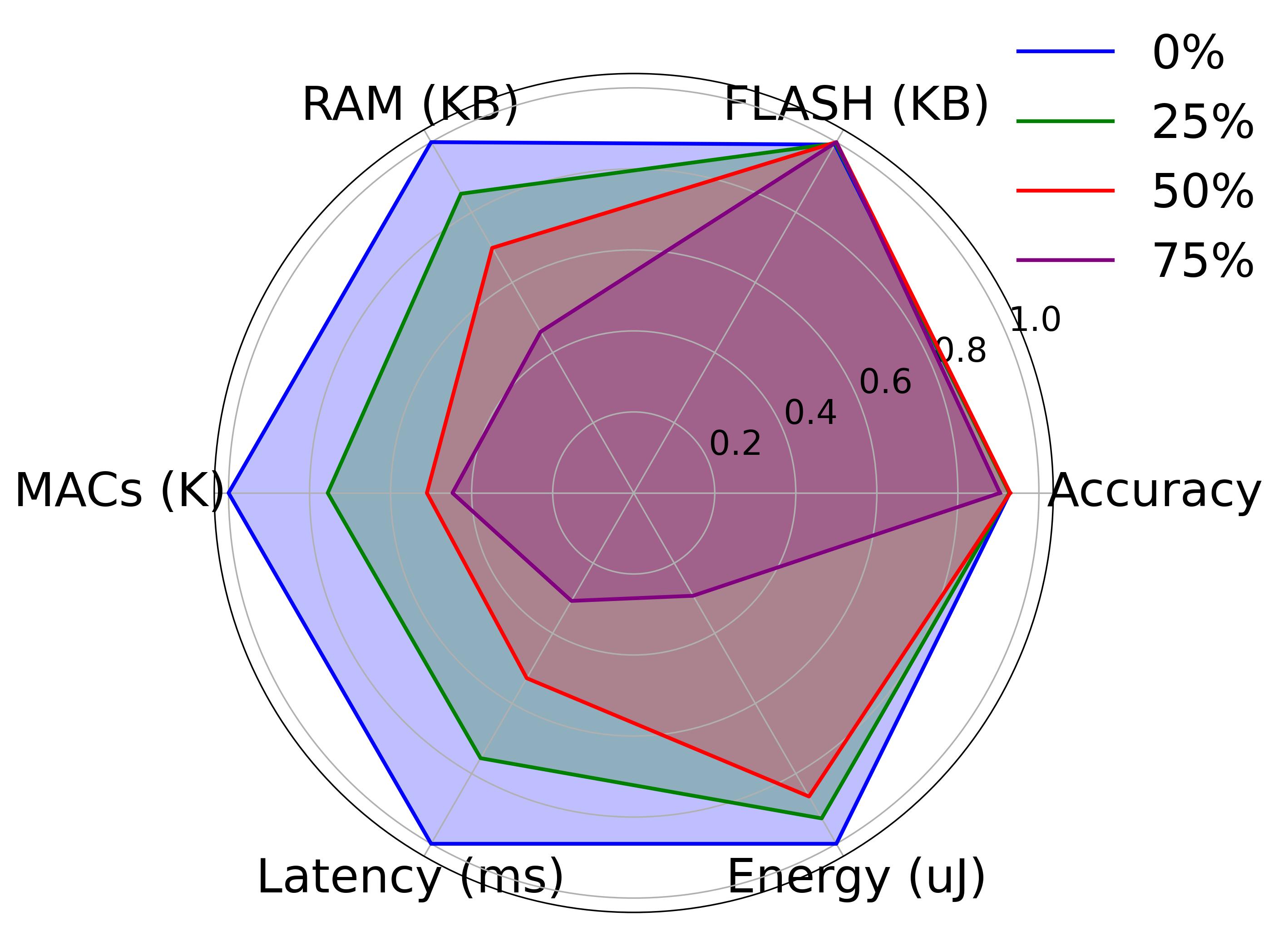}\label{fig:ucihar}}\\
\subcaptionbox{WISDM}{\includegraphics[width=0.9\linewidth]{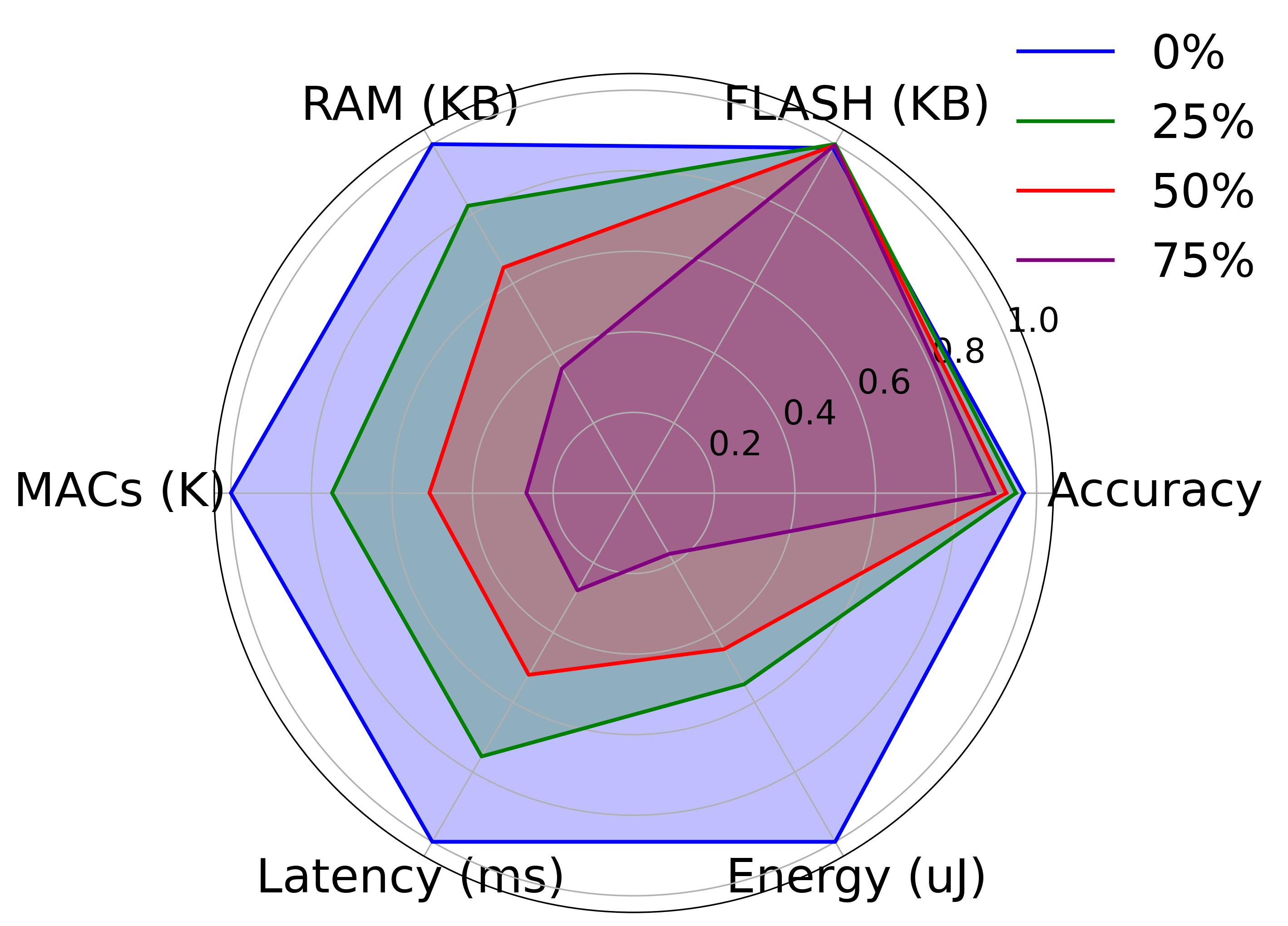}\label{fig:wisdm}}\\
\subcaptionbox{PAMAP2}{\includegraphics[width=0.9\linewidth]{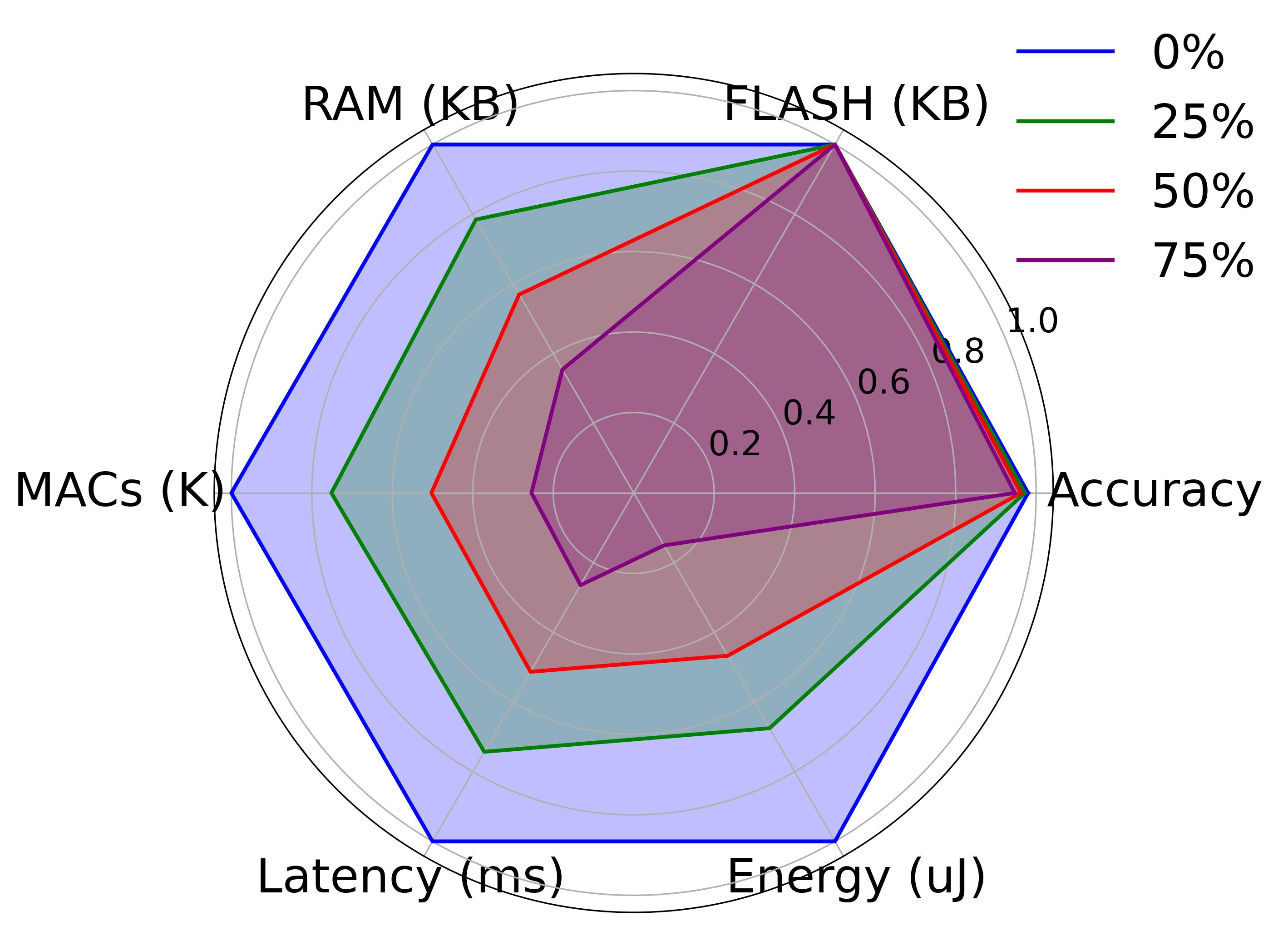}\label{fig:pamap2}}\\

\end{multicols}

\begin{multicols}{3}
\subcaptionbox{MHEALTH}{\includegraphics[width=0.9\linewidth]{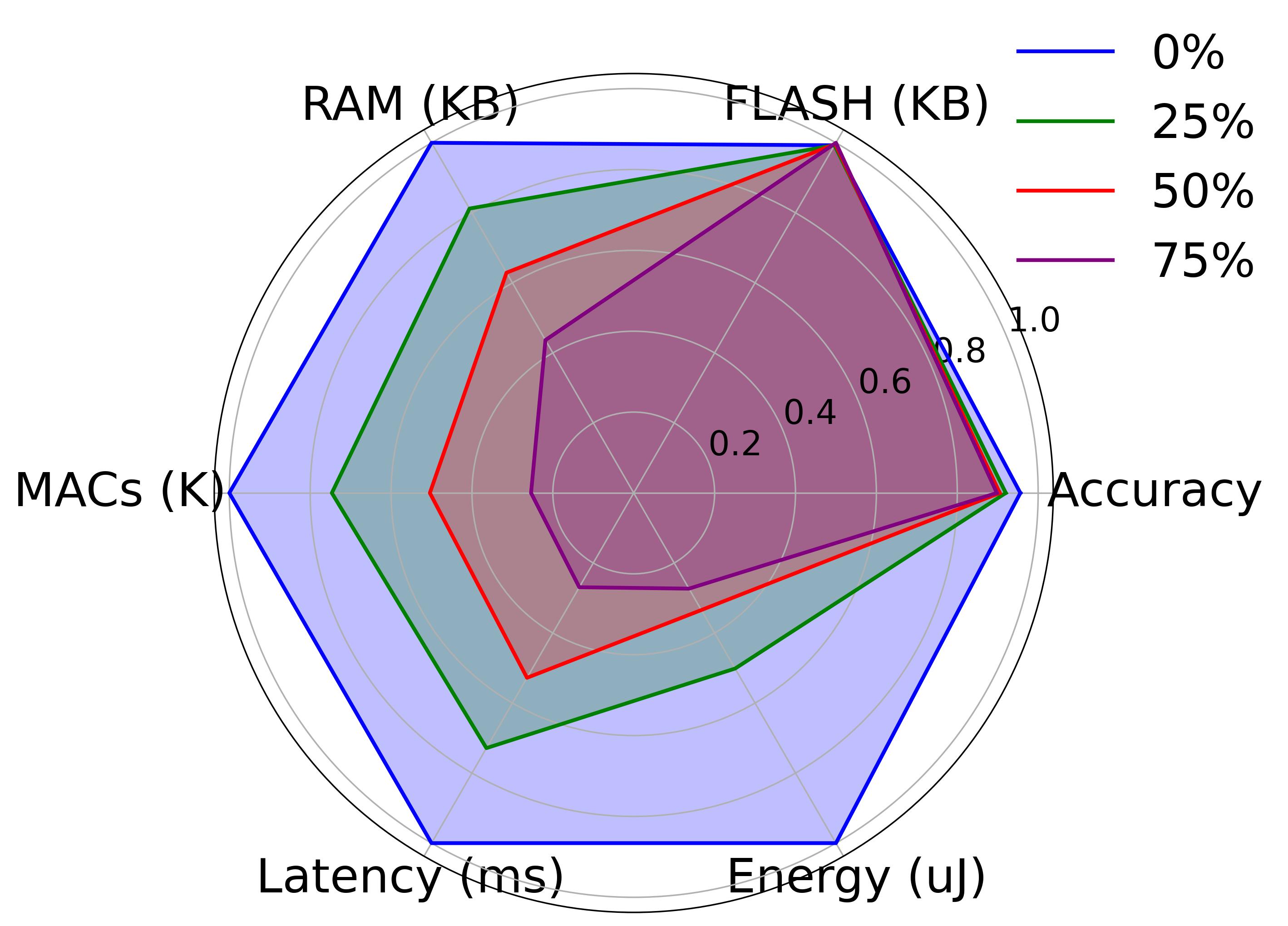}\label{fig:fig4}}\\
\hspace{-0.2in}
\subcaptionbox{MIT-BIH}{\includegraphics[width=0.9\linewidth]{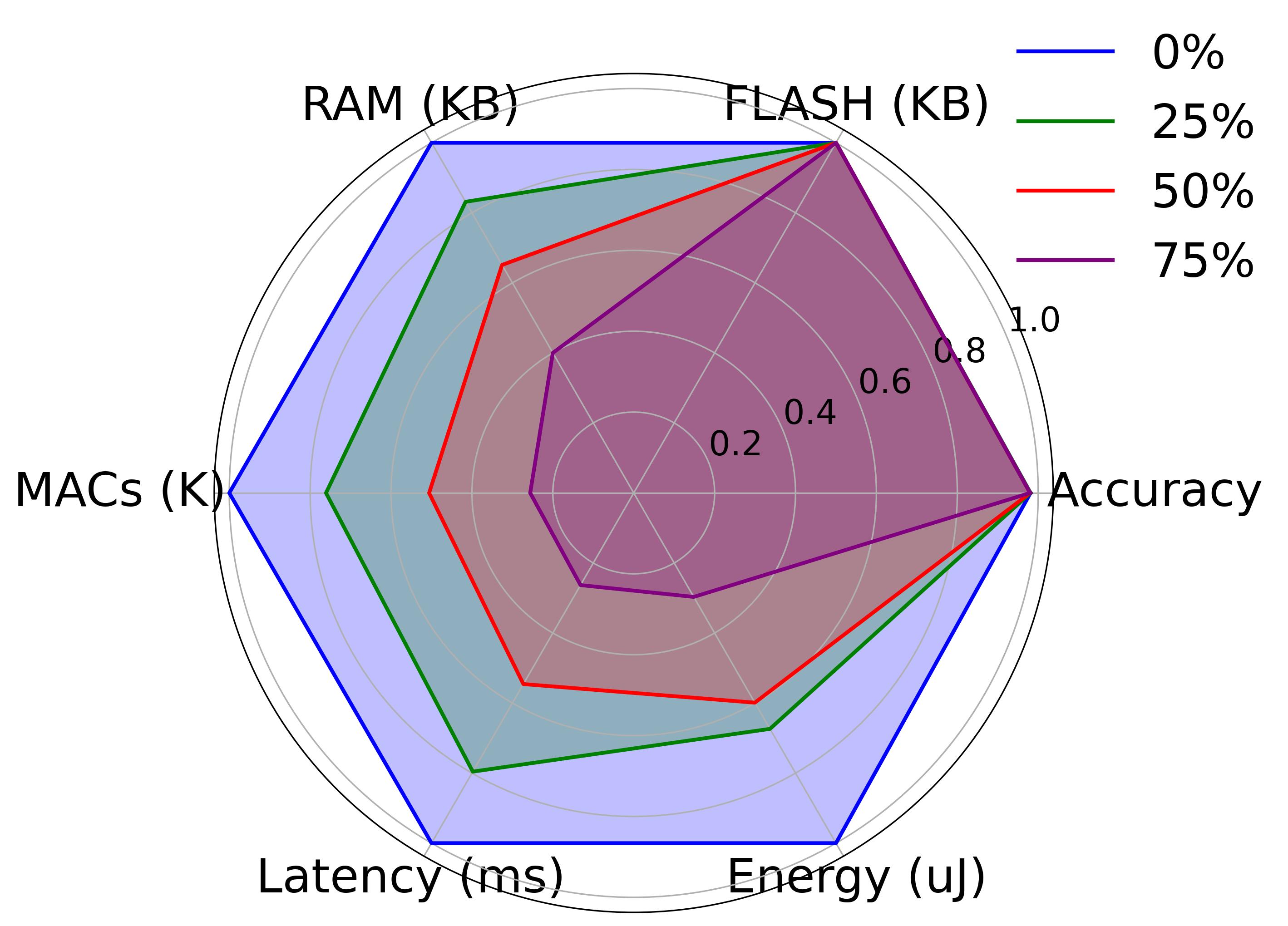}\label{fig:fig5}}\\
\subcaptionbox{PTB Diagnostic ECG Database}{\includegraphics[width=0.9\linewidth]{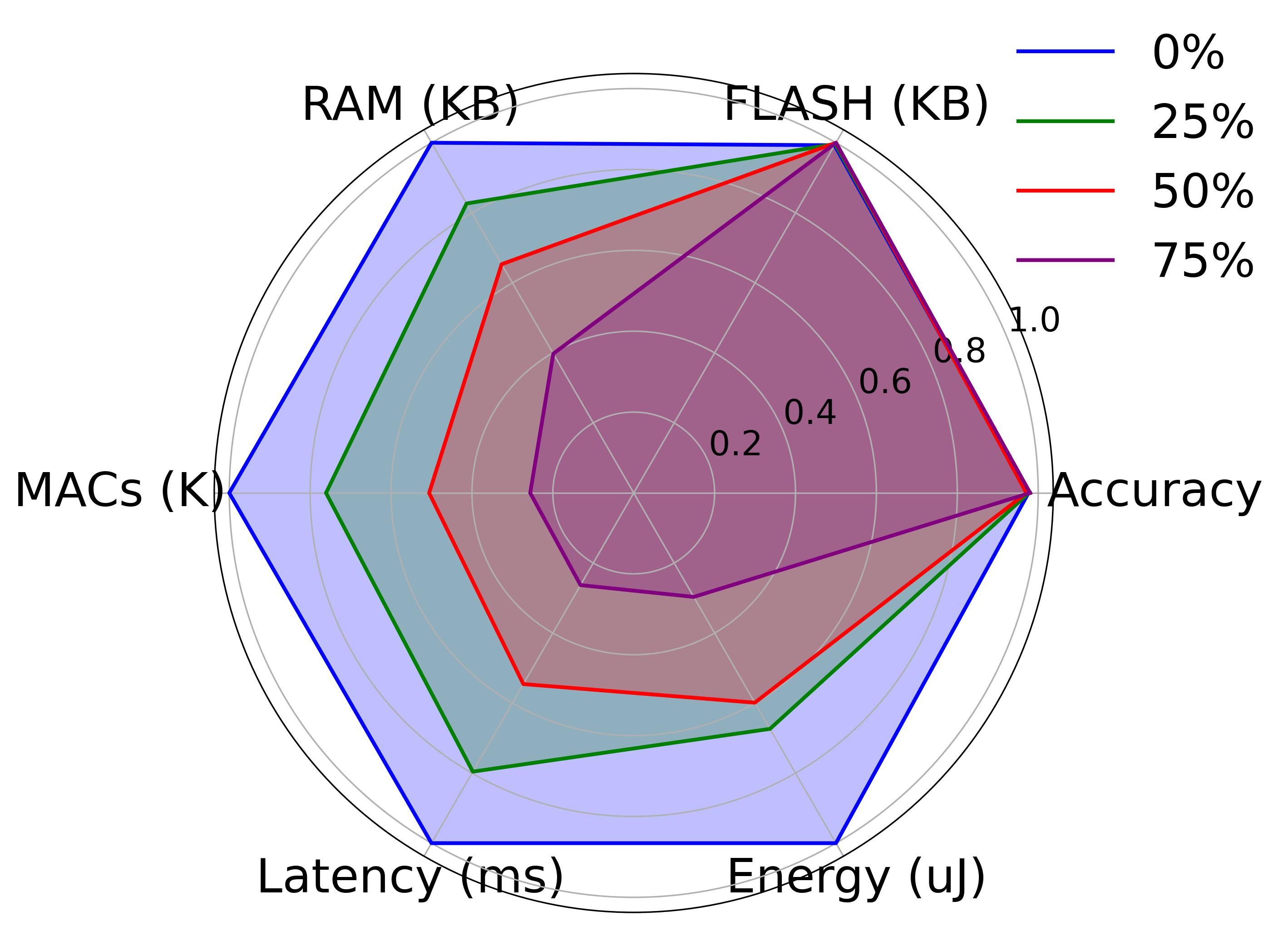}\label{fig:fig5}}\\

\end{multicols}
\vspace{-0.1in}
\caption{Radar plots comparing various metrics (Accuracy, FLASH (KB), RAM (KB), MACs (K), Time (ms), Energy (µJ)) across different data acquisition rate reduction levels: 0\% (blue), 25\% (green), 50\% (red), and 75\% (purple). Each plot is normalized for better comparison and shows performance and resource utilization trade-offs for different datasets. The shaded areas illustrate how accuracy and efficiency vary with reduction levels. This visualization highlights that significant reductions in data acquisition rate can lead to a lower resource requirement while maintaining similar levels of accuracy. }
\label{plots}
\vspace{-0.1in}
\end{figure*}

\vspace{-0.1in}
\section{Results and Discussion}
The experiments conducted on various datasets with different data rates reveal the impact on accuracy and resource footprints of optimized models in MCU environments. The results for UCIHAR, WISDM, PAMAP2, MHEALTH, MIT-BIH Arrhythmia Database, and PTB Diagnostic ECG Database are shown in Tables [ \ref{ucihar}, \ref{wisdm}, \ref{pamap2}, \ref{mhealth}, \ref{mitbih}, \ref{ptbd}], respectively. The resource footprint metrics considered include RAM, FLASH, MACs, Latency, Energy, and Accuracy. The followings are our observations:

\textbf{FLASH:} Reducing the data rate while keeping the network architecture fixed does not affect FLASH requirements, as they are independent of input shape. Optimization of data rates primarily influences dynamic memory and computational resources, while the static memory requirement (FLASH) remains constant.

\textbf{RAM :} Lowering the data rate reduces the input dimension, decreasing RAM requirements. For UCIHAR, a 50\% data rate reduction leads to a 30\% RAM reduction with negligible accuracy loss. WISDM sees a 34\% RAM reduction with a 4\% accuracy drop, PAMAP2 experiences a 45\% RAM reduction with a 2\% accuracy loss, and MHEALTH shows a 20\% RAM reduction with a 3\% accuracy drop ( at 25\% data rate reduction). MIT-BIH and PTB maintain accuracy while reducing RAM requirements by 60\% (at 75\% data rate reduction). This significant reduction in RAM usage is beneficial for deployment on resource-constrained devices.

\textbf{MACs :}
Reducing the data rate also lowers MAC requirements. For UCIHAR, a 50\% data rate reduction results in a 40\% MAC reduction with negligible accuracy loss. WISDM shows a 49\% MAC reduction with a 4\% accuracy drop, PAMAP2 achieves a 45\% MAC reduction with a 2\% accuracy loss, and MHEALTH sees a 25\% MAC reduction with a 3\% accuracy drop (at 25\% data rate reduction). MIT-BIH and PTB retain accuracy while reducing MAC requirements by 75\% (at 75\% data rate reduction). Lower data rates effectively reduce computational complexity, translating into lower power consumption and faster inference times.

\textbf{Latency :}
Reducing the data rate decreases latency. UCIHAR shows a 48\% latency reduction with a 50\% data rate reduction and minimal accuracy loss. WISDM shows a 49\% latency reduction with a 4\% accuracy drop, PAMAP2 experiences a 49\% latency reduction with a 2\% accuracy loss, and MHEALTH shows a 27\% latency reduction with a 3\% accuracy drop (at 25\% data rate reduction). MIT-BIH and PTB maintain accuracy while reducing latency requirements by 74\% (at 75\% data rate reduction). Lower latency is critical for real-time applications.

\textbf{Energy :}
Lowering the data rate reduces energy consumption. UCIHAR shows a 71\% energy reduction with a 50\% data rate reduction and minimal accuracy loss. WISDM experiences a 55\% energy reduction with a 4\% accuracy drop, PAMAP2 sees a 53\% Energy reduction with a 2\% accuracy loss, and MHEALTH achieves a 50\% energy reduction with a 3\% accuracy drop (at 25\% data rate reduction). MIT-BIH and PTB maintain accuracy while reducing energy requirements by 70\% (at 75\% data rate reduction). Energy efficiency is crucial for battery-operated IoT devices, highlighting the potential for significant improvements in battery life without substantial sacrifices in model performance.

In summary, for UCIHAR, reducing the data rate by 50\% maintains accuracy while reducing RAM by 30\%, MACs by 40\%, latency by 48\%, and energy consumption by 71\%. For WISDM, a 50\% data rate reduction results in a 4\% accuracy drop, with RAM, MACs, latency, and energy reduced by 34\%, 49\%, 49\%, and 55\%, respectively. PAMAP2 shows a 2\% accuracy loss with a 50\% data rate reduction, resulting in RAM, MACs, latency, and energy reductions of 45\%, 45\%, 49\%, and 53\%, respectively. MHEALTH, with a 25\% data rate reduction, shows a 3\% accuracy decrease, and reductions in RAM, MACs, latency, and energy by 20\%, 25\%, 27\%, and 50\%, respectively. MIT-BIH and PTB maintain accuracy with a 75\% data rate reduction, reducing RAM by 60\%, MACs by 75\%, latency by 74\%, and energy by 70\%. For tabular visualization, see Table \ref{tab:gist}.

In general, the trends observed across all datasets indicate that reducing data rates consistently lowers RAM, energy, latency, and MAC requirements. Accuracy remains unchanged in the UCIHAR, MIT-BIH, and PTB datasets, while there are minimal accuracy reductions in WISDM, PAMAP2, and MHEALTH datasets. For a graphical representation, refer to Figure \ref{plots}.

\begin{table}[!t]
\centering
\caption{Summary of how reducing data acquisition rates affects various metrics across datasets. The abbreviations : RAM Requirements Reduction (RRR), MACs Reduction (MR), Latency Reduction (LR), Energy Requirement Reduction (ERR), Accuracy Reduction (AR), Datarate Reduction (DR). }
\label{tab:gist}
\vspace{-0.1in}
\resizebox{\linewidth}{!}{%
\begin{tabular}{|l|c|c|c|c|c|c|}
\hline
\multicolumn{1}{|c|}{\textbf{Dataset}} & \textbf{\begin{tabular}[c]{@{}c@{}}RRR (\%)\end{tabular}} & \textbf{\begin{tabular}[c]{@{}c@{}} MR(\%)\end{tabular}} & \textbf{\begin{tabular}[c]{@{}c@{}}LR (\%)\end{tabular}} & \textbf{\begin{tabular}[c]{@{}c@{}}ERR (\%)\end{tabular}} & \textbf{\begin{tabular}[c]{@{}c@{}}AR (\%)\end{tabular}} & \textbf{\begin{tabular}[c]{@{}c@{}}DR (\%)\end{tabular}} \\ \hline
UCIHAR & 30 & 40 & 48 & 71 & 0 & 50 \\
WISDM & 34 & 49 & 49 & 55 & 4 & 50 \\
PAMAP2 & 45 & 45 & 49 & 53 & 2 & 50 \\
MHEALTH & 20 & 25 & 27 & 50 & 3 & 25 \\
MIT-BIH & 60 & 75 & 74 & 70 & 0 & 75 \\
PTB & 60 & 75 & 74 & 70 & 0 & 75 \\ \hline
\end{tabular}%
}
\vspace{-0.2in}
\end{table}

\vspace{-0.1in}
\section{Conclusion}
This study demonstrates that reducing data acquisition rates can significantly optimize TinyML models for deployment on resource-constrained MCUs. By lowering the sampling frequency, we achieved substantial reductions in RAM usage, MAC operations, latency, and energy consumption while maintaining near-identical classification accuracies. Our experiments across six benchmark datasets (UCI-HAR, WISDM, PAMAP2, MHEALTH, MIT-BIH Arrhythmia Database, and PTB Diagnostic ECG Database) showed that FLASH memory requirements remained constant, highlighting the effectiveness of this approach in enhancing the feasibility and efficiency of TinyML applications on low-power, battery-operated IoT devices. These findings offer a promising pathway for the practical deployment of TinyML models in various real-world scenarios.


\bibliographystyle{ACM-Reference-Format}
\bibliography{sample-base}


\begin{thebibliography}{14}


\ifx \showCODEN    \undefined \def \showCODEN     #1{\unskip}     \fi
\ifx \showDOI      \undefined \def \showDOI       #1{#1}\fi
\ifx \showISBNx    \undefined \def \showISBNx     #1{\unskip}     \fi
\ifx \showISBNxiii \undefined \def \showISBNxiii  #1{\unskip}     \fi
\ifx \showISSN     \undefined \def \showISSN      #1{\unskip}     \fi
\ifx \showLCCN     \undefined \def \showLCCN      #1{\unskip}     \fi
\ifx \shownote     \undefined \def \shownote      #1{#1}          \fi
\ifx \showarticletitle \undefined \def \showarticletitle #1{#1}   \fi
\ifx \showURL      \undefined \def \showURL       {\relax}        \fi
\providecommand\bibfield[2]{#2}
\providecommand\bibinfo[2]{#2}
\providecommand\natexlab[1]{#1}
\providecommand\showeprint[2][]{arXiv:#2}

\bibitem[Bousseljot et~al\mbox{.}(1995)]%
        {bousseljot1995nutzung}
\bibfield{author}{\bibinfo{person}{Ralf Bousseljot}, \bibinfo{person}{Dieter Kreiseler}, {and} \bibinfo{person}{Allard Schnabel}.} \bibinfo{year}{1995}\natexlab{}.
\newblock \showarticletitle{Nutzung der EKG-Signaldatenbank CARDIODAT der PTB {\"u}ber das Internet}.
\newblock  (\bibinfo{year}{1995}).
\newblock


\bibitem[David et~al\mbox{.}(2021)]%
        {david2021tensorflow}
\bibfield{author}{\bibinfo{person}{Robert David}, \bibinfo{person}{Jared Duke}, \bibinfo{person}{Advait Jain}, \bibinfo{person}{Vijay Janapa~Reddi}, \bibinfo{person}{Nat Jeffries}, \bibinfo{person}{Jian Li}, \bibinfo{person}{Nick Kreeger}, \bibinfo{person}{Ian Nappier}, \bibinfo{person}{Meghna Natraj}, \bibinfo{person}{Tiezhen Wang}, {et~al\mbox{.}}} \bibinfo{year}{2021}\natexlab{}.
\newblock \showarticletitle{Tensorflow lite micro: Embedded machine learning for tinyml systems}.
\newblock \bibinfo{journal}{\emph{Proceedings of Machine Learning and Systems}}  \bibinfo{volume}{3} (\bibinfo{year}{2021}), \bibinfo{pages}{800--811}.
\newblock


\bibitem[et. al(2014)]%
        {misc_mhealth_319}
\bibfield{author}{\bibinfo{person}{Banos~Oresti et. al}.} \bibinfo{year}{2014}\natexlab{}.
\newblock \bibinfo{title}{{MHEALTH}}.
\newblock \bibinfo{howpublished}{UCI Machine Learning Repository}.
\newblock
\newblock
\shownote{{DOI}: https://doi.org/10.24432/C5TW22}.


\bibitem[et. al(2012)]%
        {misc_human_activity_recognition_using_smartphones_240}
\bibfield{author}{\bibinfo{person}{Reyes~Ortiz et. al}.} \bibinfo{year}{2012}\natexlab{}.
\newblock \bibinfo{title}{{Human Activity Recognition Using Smartphones}}.
\newblock \bibinfo{howpublished}{UCI Machine Learning Repository}.
\newblock
\newblock
\shownote{{DOI}: https://doi.org/10.24432/C54S4K}.


\bibitem[Kachuee et~al\mbox{.}(2018)]%
        {kachuee2018ecg}
\bibfield{author}{\bibinfo{person}{Mohammad Kachuee}, \bibinfo{person}{Shayan Fazeli}, {and} \bibinfo{person}{Majid Sarrafzadeh}.} \bibinfo{year}{2018}\natexlab{}.
\newblock \showarticletitle{Ecg heartbeat classification: A deep transferable representation}. In \bibinfo{booktitle}{\emph{2018 IEEE international conference on healthcare informatics (ICHI)}}. IEEE, \bibinfo{pages}{443--444}.
\newblock


\bibitem[Labs(2024)]%
        {mltk}
\bibfield{author}{\bibinfo{person}{Silicon Labs}.} \bibinfo{year}{2024}\natexlab{}.
\newblock \bibinfo{title}{{MLTK}: Machine Learning Toolkit}.
\newblock \bibinfo{howpublished}{\url{https://siliconlabs.github.io/mltk/}}.
\newblock
\newblock
\shownote{Accessed: 2024-07-03}.


\bibitem[Moody and Mark(2001)]%
        {moody2001impact}
\bibfield{author}{\bibinfo{person}{George~B Moody} {and} \bibinfo{person}{Roger~G Mark}.} \bibinfo{year}{2001}\natexlab{}.
\newblock \showarticletitle{The impact of the MIT-BIH arrhythmia database}.
\newblock \bibinfo{journal}{\emph{IEEE engineering in medicine and biology magazine}} \bibinfo{volume}{20}, \bibinfo{number}{3} (\bibinfo{year}{2001}), \bibinfo{pages}{45--50}.
\newblock


\bibitem[Reiss(2012)]%
        {misc_pamap2_physical_activity_monitoring_231}
\bibfield{author}{\bibinfo{person}{Attila Reiss}.} \bibinfo{year}{2012}\natexlab{}.
\newblock \bibinfo{title}{{PAMAP2 Physical Activity Monitoring}}.
\newblock \bibinfo{howpublished}{UCI Machine Learning Repository}.
\newblock
\newblock
\shownote{{DOI}: https://doi.org/10.24432/C5NW2H}.


\bibitem[Saha et~al\mbox{.}(2023a)]%
        {saha2023bandx}
\bibfield{author}{\bibinfo{person}{Bidyut Saha}, \bibinfo{person}{Riya Samanta}, \bibinfo{person}{Soumya Ghosh}, {and} \bibinfo{person}{Ram~Babu Roy}.} \bibinfo{year}{2023}\natexlab{a}.
\newblock \showarticletitle{BandX: An intelligent IoT-band for human activity recognition based on TinyML}. In \bibinfo{booktitle}{\emph{Proceedings of the 24th International Conference on Distributed Computing and Networking}}. \bibinfo{pages}{284--285}.
\newblock


\bibitem[Saha et~al\mbox{.}(2023b)]%
        {saha2023wrist}
\bibfield{author}{\bibinfo{person}{Bidyut Saha}, \bibinfo{person}{Riya Samanta}, \bibinfo{person}{Soumya~Kanti Ghosh}, {and} \bibinfo{person}{Ram~Babu Roy}.} \bibinfo{year}{2023}\natexlab{b}.
\newblock \showarticletitle{From Wrist to World: Harnessing Wearable IMU Sensors and TinyML to Enable Smart Environment Interactions}. In \bibinfo{booktitle}{\emph{Proceedings of the Third International Conference on AI-ML Systems}}. \bibinfo{pages}{1--3}.
\newblock


\bibitem[Saha et~al\mbox{.}(2023c)]%
        {saha2023tinyml}
\bibfield{author}{\bibinfo{person}{Bidyut Saha}, \bibinfo{person}{Riya Samanta}, \bibinfo{person}{Soumya~Kanti Ghosh}, {and} \bibinfo{person}{Ram~Babu Roy}.} \bibinfo{year}{2023}\natexlab{c}.
\newblock \showarticletitle{TinyML-Driven On-Device Personalized Human Activity Recognition and Auto-Deployment to Smart Bands}. In \bibinfo{booktitle}{\emph{Proceedings of the Third International Conference on AI-ML Systems}}. \bibinfo{pages}{1--9}.
\newblock


\bibitem[Saha et~al\mbox{.}(2024)]%
        {saha2024personalized}
\bibfield{author}{\bibinfo{person}{Bidyut Saha}, \bibinfo{person}{Riya Samanta}, \bibinfo{person}{Ram~Babu Roy}, \bibinfo{person}{Chinmay Chakraborty}, {and} \bibinfo{person}{Soumya~K Ghosh}.} \bibinfo{year}{2024}\natexlab{}.
\newblock \showarticletitle{Personalized Human Activity Recognition: Real-time On-device Training and Inference}.
\newblock \bibinfo{journal}{\emph{IEEE Consumer Electronics Magazine}} (\bibinfo{year}{2024}).
\newblock


\bibitem[{TensorFlow Lite}(2024)]%
        {tensorflowlite}
\bibfield{author}{\bibinfo{person}{{TensorFlow Lite}}.} \bibinfo{year}{Accessed on July 5, 2024}\natexlab{}.
\newblock \bibinfo{title}{TensorFlow Lite}.
\newblock
\newblock
\urldef\tempurl%
\url{https://www.tensorflow.org/lite}
\showURL{%
\tempurl}


\bibitem[Weiss(2019)]%
        {misc_wisdm_smartphone_and_smartwatch_activity_and_biometrics_dataset__507}
\bibfield{author}{\bibinfo{person}{Gary Weiss}.} \bibinfo{year}{2019}\natexlab{}.
\newblock \bibinfo{title}{{WISDM Smartphone and Smartwatch Activity and Biometrics Dataset }}.
\newblock \bibinfo{howpublished}{UCI Machine Learning Repository}.
\newblock
\newblock
\shownote{{DOI}: https://doi.org/10.24432/C5HK59}.


\end{thebibliography}

\end{document}